\documentclass[conference]{IEEEtran}
\IEEEoverridecommandlockouts

\usepackage{cite}
\usepackage{xurl}
\usepackage{amsmath,amssymb,amsfonts}
\usepackage{algorithmic}
\usepackage{graphicx}
\usepackage{textcomp}
\usepackage{xcolor}
\usepackage{algorithm}
\usepackage{algorithmic}
\def\BibTeX{{\rm B\kern-.05em{\sc i\kern-.025em b}\kern-.08em
    T\kern-.1667em\lower.7ex\hbox{E}\kern-.125emX}}
\begin{document}

\title{Centralized and Federated Heart Disease Classification Models Using UCI Dataset and their Shapley-value Based Intepretability}

\author{\IEEEauthorblockN{Mario Padilla Rodriguez}
\IEEEauthorblockA{\textit{Electrical and Computer Engineering} \\
\textit{University of Detroit Mercy}\\
padillma1@udmercy.edu}
\and
\IEEEauthorblockN{Mohamed Nafea}
\IEEEauthorblockA{\textit{Electrical and Computer Engineering} \\
\textit{University of Detroit Mercy}\\
nafeamo@udmercy.edu}
}

\maketitle

\begin{abstract}
Cardiovascular diseases are a leading cause of mortality worldwide, highlighting the need for accurate diagnostic methods. This study benchmarks centralized and federated machine learning  algorithms for heart disease classification using the UCI  dataset which includes 920 patient records from four hospitals in the USA, Hungary and Switzerland. Our benchmark is supported by Shapley-value interpretability analysis to quantify features' importance for classification. In the centralized setup, various binary classification algorithms are trained on pooled data, with a support vector machine (SVM) achieving the highest testing accuracy of 83.3\%, surpassing the established benchmark of 78.7\% with logistic regression. Additionally, federated learning algorithms with four clients (hospitals) are explored, leveraging the dataset's natural partition to enhance privacy without sacrificing accuracy. Federated SVM, an uncommon approach in the literature, achieves a top testing accuracy of 73.8\%. Our interpretability analysis aligns with existing medical knowledge of heart disease indicators. Overall, this study establishes a benchmark for efficient and interpretable pre-screening tools for heart disease while maintaining patients' privacy.

 
\end{abstract}


\section{Introduction and Background}
\subsection{Motivation}
Cardiovascular diseases remain a leading global cause of death highlighting the urgent need for accurate and efficient diagnostic tools. Over the past decade, machine learning (ML) has shown promise in revolutionizing healthcare diagnostics. This paper offers a comprehensive benchmark study of centralized and federated ML algorithms for heart disease classification using the UCI heart disease dataset \cite{UCI_heart}. Due to the presence of several inflated results in the literature based on this dataset, our benchmark provides a useful reference for further development. Additionally, our benchmark is supported by Shapley-value based interpretability analysis to quantify importance of features in influencing the decision, enhancing the transparency and trustworthiness of the adopted predictive models for healthcare professionals. In the following, we outline our contributions and provide background on federated learning and model interpretability.

\subsection{Background}
\paragraph*{Federated learning} In privacy-sensitive applications such as healthcare, there exist strict privacy guidelines protecting patient's information \cite{puhan2023digital}. 
To comply with these critical guidelines, cloud-center (centralized) training of ML models in healthcare applications involve strong anonymization or encryption of patients data, for example using metrics such as k-anonymity \cite{sweeney2002k}, strictly limiting their utility. Such limitations resulted in the recent popularity of federated learning (FL), a privacy-preserving, communication and computation efficient, distributed learning paradigm, in healthcare \cite{rieke2020future,xu2021federated,nguyen2022federated, ogier2022flamby}. FL enables several clients holding sensitive data to collaboratively train ML models for better generalization, while promoting higher privacy guarantees via maintaining local data local, see \cite{Konecny2016federated,mcmahan2017communication,kairouz2021advances}. Instead, clients train local model updates and only these updates are shared and aggregated globally with a coordinating central server in an iterative fashion. These updates are often ephemeral (deleted after use) and contain significantly less (private) information than raw or anonymized data, hence enhancing privacy protection of clients data, and improving communication efficiency and computational requirements at the server. 

FL however is typically characterized by high inter-client statistical dataset heterogeneity \cite{du2021collaborative,li2020using,pati2021federated}. To cope with data heterogeneity, researchers proposed several methods for improving local model performance, while continue to benefit from improved generalization due to global training. These include (i) improving worst-case performance of the global model by ensuring uniform performance across clients \cite{karimireddy2020scaffold,li2020federated,reddi2020adaptive}, and (ii) training personalized local models, for example by fine tuning the global model on local clients' data or regularizing local models with global penalties \cite{cho2021personalized,fallah2020personalized,ghosh2020efficient,jiang2019improving,nafea2022proportional,sattler2019clustered,smith2017federated}.

In order to examine the FL algorithms that address data heterogeneity, researches tend to synthetically simulate data partitioning across clients. Common approaches to produce such synthetic partitions for classification datasets include associating samples from a limited number of classes to each client \cite{mcmahan2017communication} and Dirichlet sampling based on the class labels \cite{hsu2019measuring}. The UCI heart disease dataset provides a great example of naturally heterogeneous splits of the data across geographically distant hospitals. Therefore, providing a benchmark for FL on this dataset is of an active interest. 

\paragraph*{Interpretability} Recently, there has been a growing interest in transparency and interpretability analysis of machine learning systems \cite{datta2016algorithmic, doshi2017towards, arrieta2020explainable}. Interpretable ML aims to clarify the often opaque inner workings of black-box models, enhancing understanding of the features influencing their decisions. In healthcare applications, understanding how a ML model produces its predictions could (i) provide critical insights into disease mechanisms; (ii) guide clinical decision-making; (iii) increase clinician acceptance and adoption by aligning with medical expertise and ethical standards, and overall (iv) improve patient outcomes. In this work, we integrate interpretability analysis into our benchmark to facilitate their practical application in clinical practice. 

\paragraph*{Shapley-value based interpetability} 
The Shapley value function, derived from cooperative game theory, quantifies the marginal contributions of individuals to the overall utility generated by a coalition of participants \cite{shapley1953value}. It does this by evaluating and averaging the contributions of an individual participant to the overall utility when considered with any possible coalition of other participants. In ML, they are used to attribute a model's predictions to its input features by assessing each feature's marginal impact on prediction accuracy. Shapley values support a theoretically sound and easily understandable interpretability analysis. They are desirable due to their key properties: (i) Efficiency, where the sum of Shapley values for all features equals the total utility (accuracy) of the model's predictions; (ii) Symmetry, meaning features with identical contributions across all coalitions receive equal Shapley values; (iii) Dummy features (with no contribution to prediction) receive a Shapley value of zero; and (iv) Monotonicity, ensuring that if feature $i$ consistently contributes more to the overall utility than feature $j$ across all possible coalitions, then feature $i$ has a higher Shapley value. \cite{lundberg2017unified} extends Shapley values to the SHAP (SHapley Additive exPlanations) framework which systematically decomposes model predictions into additive contributions from each feature. SHAP also includes efficient approximations for Shapley values in large-scale models and provides a unified framework for interpreting various types of ML models, including tree-based models and deep neural networks. In this work, we utilize the SHAP framework for integrating interpretability results to our benchmark. 

\subsection{Contributions}
\paragraph*{Centralized benchmark} We train various binary classification algorithms, including logistic regression (LR), neural networks (NN), support vector machines (SVM), k-nearest neighbor (KNN), Naïve Bayes (NB), and random forests. Notably, we achieve centralized accuracy of 83.3\% via the application of a linear-kernel SVM model, surpassing established standards such as the 78.7\% accuracy attained by the logistic regression (LR) model in \cite{ogier2022flamby}. 

\paragraph*{Federated benchmark} Our federated benchmark for the UCI heart disease dataset improves over the benchmark reported in \cite{ogier2022flamby}. In particular, we use the top three performing models in the centralized setup as local trainers for the federated setting. We examine various federated aggregation strategies including FedAvg \cite{mcmahan2017communication}, FedAdam \cite{reddi2020adaptive}, FedYogi \cite{reddi2020adaptive}, and SCAFFOLD \cite{karimireddy2020scaffold}. Once again, SVM is the top-performing model in the federated setting, achieving the highest test accuracy across the four aggregation strategies; the highest being 73.8 \% using FedYogi. Federating SVMs is not common in the literature; there exist only a handful works that uses federated SVMs which are all outside our scope of applications. To meet our specific needs, we develop our own version of federated SVM from scratch. Finally, except for the 1-layer NN model, FedAvg seems to perform closely to the more complicated aggregation algorithms like FedAdam and FedYogi.  

\paragraph*{Interpretability Analysis}
Computing SHAP values for the centralized models highlights four features that are most important factors for predictions. The first is the ST depression induced by exercise (`oldpeak'): The ST segment is the part of the ECG between the end of the S wave and start of the T wave \cite{ncbi2024ST}. Patients with ST depression are up to 2.34 times more likely to suffer from heart failure than those without \cite{kawaji2023clinical}. Another important feature is the type of chest pain (`cp’): Out of its four categories, typical angina is a well-known symptom of heart conditions, most notably coronary artery disease \cite{nhlbi2024ST}. Another identified feature is 'exang', indicating whether exercise-induced angina is present. Similar to 'cp', angina is a common precursor to heart disease.

The rest of this paper is organized as follows. Section \ref{related} discusses the related work. Section 3 presents our methodology for tuning and testing the models, as well as the results obtained for the centralized and federated settings. Section 4 discusses the interpretability of the models, highlighting the features that are most and least important to the predictions. Section 5 concludes the paper and discusses future work. 




\section{Related Work} \label{related}

A substantial body of literature explores ML models for heart disease detection. \cite{ahsan2022machine,aljanabi2018machine,azmi2022systematic} detail various methods and algorithms with binary classification being the most common approach. Some studies use altered versions of the Cleveland dataset from the UCI collection, which include duplicated entries \cite{kaggle_heart}. For example, the dataset is expanded from the original 303 entries to 1025 by adding over 700 duplicates. Validation on such duplicated data can lead to unreliable and inflated performance metrics. A recent study \cite{prasher2023evaluation} reports a 99.7\% accuracy using a KNN model on this duplicated dataset. Other examples of studies using this data include \cite{ali2021heart}, \cite{ahmad2022efficient}, and \cite{ahamad2023influence}. In contrast, our work provides a reliable benchmark using the unmodified UCI heart disease datasets.

FLamby \cite{ogier2022flamby} is a recent benchmark suite for cross-silo federated learning (FL) in healthcare. It addresses the need for realistic datasets with natural partitions suitable for cross-silo FL settings involving a few reliable clients with medium to large datasets, typically found in healthcare and finance applications. The FLamby suite includes $7$ healthcare datasets with diverse tasks, data modalities, and volumes, and provides baseline training code along with benchmarks for various FL aggregation algorithms. Our work enhances the benchmark for the UCI heart disease dataset included in FLamby. We find that SVM outperforms logistic regression, utilized by Flamby for this dataset, in both centralized and federated settings

Several works incorporate interpretability in heart disease classification.  \cite{ayano2022interpretable} reviews interpretable ML methods using electrocardiogram (ECG) data. Processing time-series data like ECGs requires complex models, making their decisions harder to understand. \cite{anand2022explainable} uses SHAP to highlight ECG segments most relevant for model prediction. \cite{goodfellow2018towards} employs class activation maps (CAM) \cite{zhou2016learning} to visualize the ECG regions a convolutional neural network (CNN) emphasizes. Interpretability analysis has also been applied to models using tabular data. \cite{wang2021interpretable} trained six models on a private dataset including patient demographics, medical history, physical status, vitals, medical therapy, echocardiography, electrocardiography, and lab parameters. They used SHAP to predict mortality in heart failure patients over a 3-year period and identify influential features. \cite{sethi2024explainable} visualized SHAP values for models trained on augmented versions of the UCI database to create a user-friendly prediction interface without analyzing important features. In contrast, our work is the first to perform interpretability analysis on the unmodified UCI heart disease database, enhancing the practicality of our established benchmark.



\section{Experimental setup}
\subsection{Dataset details}
The dataset consists of 920 patient records spread across four separate hospitals' databases in Cleveland, Hungary, Switzerland, and the VA Long Beach. The Cleveland dataset is the most commonly known of the four UCI datasets. It consists of data collected from 303 patients who were referred for coronary angiography at the Cleveland Clinic from May 1981 to September 1984. These patients had no previous history or electrocardiographic signs of heart disease. All 303 patients underwent a thorough evaluation including medical history assessment, physical examination, resting electrocardiogram, serum cholesterol testing, and fasting blood sugar measurement as part of their routine assessment. 

For the Long Beach VA dataset, the patients included all individuals who underwent cardiac catheterization at the Long Beach Veterans Administration Medical Center between 1984 and 1987. After excluding those with prior infarction, valvular disease and prior catheterization, there were 200 patient data points remaining in the dataset. Disease prevalence is 75\%.

The Hungary dataset is comprised of patients who underwent catheterization at the Hungarian Institute of Cardiology in Budapest from 1983 to 1987. Patients with prior heart attacks or valvular diseases were not included, resulting in 425 subjects, and reduced further to 294 after data preprocessing. Disease prevalence is 38\%. 

For the Switzerland dataset, data was collected from individuals who underwent cardiac catheterization at the university hospitals in Zurich and Basel, Switzerland, in 1985 (excluded patients similar to the Long Beach VA dataset). Out of the 143 Swiss patients, 58 had the procedure done in Zurich, and 85 in Basel. Data preprocesssing resulted in 123 Swiss patient records. Disease prevalence 84\%.

There are 76 total attributes, but all research conducted on the dataset used only the 14 features indicated in Table \ref{tab:UCI}.
\begin{table}[!t]
\centering
\caption{ }
\begin{tabular}{|c|l|l|}
\hline
\textbf{Feature Number} & \textbf{Feature} & \textbf{Details} \\
\hline
3 & age       & age in years \\
\hline
4 & sex       & 1=male, 0=female \\
\hline
9 & cp        & chest pain type \\
\hline
10 & trestbps  & resting blood pressure \\
\hline
12 & chol      & serum cholesterol \\
\hline
16 & fbs       & fasting blood sugar \\
\hline
19 & restecg   & resting ECG results \\
\hline
32 & thalach   & max heart rate \\
\hline
38 & exang     & exercise induced angina \\
\hline
40 & oldpeak   & ST depression due to exercise \\
\hline
41 & slope     & slope of exercise ST segment \\
\hline
44 & ca        & \# of major vessels \\
\hline
51 & thal      & Thalassemia disorder presence \\
\hline
58 & num       & the predicted attribute \\
\hline
\end{tabular}
\label{tab:UCI}
\end{table}

\subsection{Centralized binary classification models}
Binary classification models are popular in heart disease detection and other healthcare applications. We train various binary classification models on the combined (pooled) data, including logistic regression (LR); fully-connected neural network (NN) with one hidden layer; support vector machine (SVM), k-nearest neighbor (KNN); Naïve Bayes (NB); descision tree (DT), and random forests (RF).

We first train the logistic regression model in \cite{ogier2022flamby} with a learning rate of 0.001, batch size of 4, and for 30 epochs. For other models and hyperparameter combinations, each model is trained on 10 different seeds for data splits and average accuracy is computed. Hyperparameters tuned include learning rate ($0.001 - 0.1$), batch size ($4 - 64$), hidden units ($4 - 16$) for one-hidden-layer NN, SVM regularization strength $C$ ($0.01 - 10$), Laplace smoothing $\alpha$ ($0.1 - 10$) for non-Gaussian Naive Bayes, Decision tree max depth (none $- 10$), number of RF estimators ($100 - 500$), and $k$-nearest neighbors ($1 - 10$). For a fair comparison, we follow the data pre-processing steps from \cite{ogier2022flamby}, including removing entries with missing values (reducing the dataset to 740 points), using a 66\%-34\% train-test split, converting labels ``$1-4$" to "$1$" for binary classification (0 indicating no heart disease and 1 for heart disease), and normalizing the dataset. Initially, we focus on pooled performance (no federation). The top three models will be tested in the federated setting. Algorithm 1 outlines the testing process for centralized data, while Algorithm 2 covers the federated setting. 

\subsection{Federated aggregation strategies}
Federated learning (FL)  reduces communication overhead compared to traditional centralized approaches by exchanging model updates, in an iterative fashion, instead of raw data. FL also preserves data privacy and reduces bandwidth usage, crucial for large-scale and distributed environments. In it's general form, it begins each round by sampling a subset of clients $\mathcal{S}$ from the existing number of $N$ clients. Each client $i \in \mathcal{S}$ performs local training on its own data in parallel with other selected clients. After local updates, clients communicate their model updates to the central server. At the central server, the received model updates are aggregated using one of the aggregation schemes to be explained in this section. In this work, all of the four clients are selected in each federation round, i.e., no client sampling. 

The benchmark in \cite{ogier2022flamby} provides a code to several federated aggregation algorithms, including FedAvg, FedAdam, Scaffold, and FedProx. In the following, we provide the technical details for each of these aggregation schemes.

\textbf{FedAvg} \cite{mcmahan2017communication} is the first federated aggregation algorithm introduced in the literature. It operates through iterative rounds of local training, where each round involves clients performing local updates on mini-batches, followed by a central server aggregating the local models. The aggregation weights depend on the number of data points at each client. Suppose we have $K$ clients, where client $k$ has $n_k$ data points, and the total number of data points is $n$, then the aggregated model at any federation round $t$ is given by
\[\theta^{(t)} = \sum_{k=1}^K \frac{n_k}{n} \theta_k^{(t)} \]
where $\theta_k^{(t)}$ are the trained model parameters at client $k$ and federation round $t$. That is, the received model updates are aggregated using weighted averaging, where the weights are determined by the proportion of data each client holds relative to the total dataset size. This ensures that clients with more data contribute more significantly to the global model, maintaining fairness in representation. In this work, as in \cite{ogier2022flamby}, local updates in batches are counted rather than local epochs, aligning with theoretical formulations and extending to related strategies based on FedAvg. 

Despite its widespread adoption, FedAvg has several notable weaknesses. Firstly, FedAvg can suffer from high communication costs, especially when dealing with a large number of participating devices or clients. Each round of training requires the transmission of model updates between clients and the central server, which can become a bottleneck in environments with limited bandwidth or high latency. Secondly, FedAvg often struggles with non-IID (non-Independent and Identically Distributed) data, where the data distribution across clients is heterogeneous. This can lead to biased updates and a global model that may not perform well across all clients. This sensitivity to heterogeneity gets worse as the number of communication rounds increases. It is also slow to converge due to the aggregation stage, where it may struggle to average updates from clients with unique data distributions. The following algorithms often outperform FedAvg in the case of high data heterogeneity. 

\begin{algorithm}[t!]
\caption{Local classifier}
\label{alg:non_federated_learning}
\begin{algorithmic}[1]
\REQUIRE Dataset $D$ (920 points)
\STATE Set hyperparameter values for chosen model
\STATE Remove points with missing values from $D$
\STATE Split $D$ into 66\% training set $T$ and 34\% test set $E$
\STATE Normalize $T$ (e.g., scale features to have zero mean and unit variance)
\STATE Initialize model parameters $\theta$
\FOR{$t=1$ to 50 epochs}
    \STATE Update model parameters $\theta$ by training on $T$
\ENDFOR
\STATE Evaluate final model on test set $E$
\STATE Compute testing accuracy $Acc$
\STATE Output $Acc$ as the final testing accuracy
\end{algorithmic}
\end{algorithm}

\begin{algorithm}[!t]
\caption{Federated aggregation}
\label{alg:federated_learning}
\begin{algorithmic}[1]
\REQUIRE Dataset $D$ from 4 clients ($D_1, D_2, D_3, D_4$)
\STATE Set hyperparameter values for chosen model
\STATE Initialize global model parameters $\theta$
\FOR{each client $k \in \{1, 2, 3, 4\}$}
    \STATE Preprocess data similar to Local Classifier
    \FOR{$t=1$ to 50 epochs}
        \STATE Local Classifier 
        \STATE Send $\theta_k$ to the server
        \STATE Receive updated global model parameters $\theta$ from the server
    \ENDFOR
    \STATE Evaluate local model on test set $E_k$
    \STATE Compute local testing accuracy $Acc_k$
\ENDFOR
\STATE Aggregate local testing accuracies $Acc_1, Acc_2, Acc_3, Acc_4$
\STATE Compute average testing accuracy $\text{Acc}_{\text{avg}} = \sum_{k=1}^{4} \frac{n_k}{n} Acc_k$
\STATE Output $\text{Acc}_{\text{avg}}$ as the final testing accuracy
\end{algorithmic}
\end{algorithm}

\textbf{FedAdam} \cite{reddi2020adaptive} is an extension of the FedAvg algorithm that incorporates the ADAM \cite{kingma2014adam} optimizer for improved performance. It combines FedAvg's model aggregation with ADAM's adaptive learning rate mechanism, allowing FedAdam to handle non-stationary and sparse gradients typically encountered in FL settings. By dynamically adjusting learning rates based on the magnitude and direction of gradients, ADAM optimizes the training process for each client individually, contributing to more accurate model updates. Client updates are computed as:
\vspace{-0.4em}
\begin{equation}
\begin{aligned}
    \Delta_t &= \frac{1}{|S|} \sum_{i \in S} \Delta_i^t \\
    m_{t} &= \beta_1 m_{t-1} + (1 - \beta_1) \Delta_t \\
    v_{t} &= \beta_2 v_{t-1} + (1 - \beta_2) (\Delta_t)^2 \\
    x_{t+1} &= x_{t} - \eta \frac{{m}_{t}}{\sqrt{{v}_{t}} + \tau}
\end{aligned}
\label{eq:FedAdam_updates}
\end{equation}
where $\Delta_t$ is the aggregated gradient update at time $t$, $S$ is the set of clients participating, $|S|$ is the number of clients in $S$, $m_t$ and $v_t$ are the first and second moment estimates, $\beta_1$ and $\beta_2$ are the exponential decay rates for these moment estimates, $\eta$ is the learning rate, and $\tau$ is a small constant for numerical stability. 

Each client then sends these gradients to the server, which aggregates them and updates the global model accordingly. This approach helps address issues related to heterogeneous data and varying computation capabilities among clients, leading to more efficient and effective federated learning (FL) training. Specifically, by employing the ADAM optimizer, FedAdam optimizes learning rates based on local gradients, which helps mitigate the impact of data distribution variations across clients. This adaptive approach results in more robust and efficient FL training, improving model convergence and overall performance metrics compared to FedAvg.

\textbf{FedYogi} \cite{reddi2020adaptive} is designed to be a more adaptive and efficient variant of FedAvg, leveraging techniques to handle non-IID data distributions and optimize communication for improved federated aggregation. It uses adaptive sampling techniques to select and prioritize client updates based on factors such as data distribution characteristics and model performance. Specifically, to address the issue of heterogeneity, FedYogi dynamically adjusts the learning rate based on the gradients' first and second moments, $m_t$ and $v_t$, respectively. For each client, the updates are computed similarly to FedAdam, except for the second moment estimate, $v_t$, which is computed as
\vspace{-0.4em}
\begin{align*}
v_{t} &= v_{t-1} - (1 - \beta_2)\Delta_t^2\text{sign}(v_{t-1} - \Delta_t^2)
\end{align*}
where $v_t, \beta_2, \Delta_t$ are the same as in \eqref{eq:FedAdam_updates}. The difference in how $v_t$ is computed between the two strategies is that when $v_{t-1}$ is much larger than $\Delta_t^2$, both strategies will rapidly increase the learning rate, but FedYogi does so in a more controlled way \cite{zaheer2018adaptive}.

\textbf{SCAFFOLD} \cite{karimireddy2020scaffold} tackles client drift in FL by using control variables. Client drift occurs when individual clients' models diverge from the global model due to differences in data, computation capabilities, or environmental changes, leading to inconsistencies and reduced overall performance. \cite{ogier2022flamby} enhances this approach to ensure full client participation while minimizing data exchange between clients and the server. SCAFFOLD uses control variables to adjust local model updates based on discrepancies with the global model. Specifically, each client has a control variable $\boldsymbol{c_i}$, and the server maintains a global control variable $\boldsymbol{c}$. Updates for the global and local control variables and model parameters are computed as follows:
\vspace{-0.4em}
\begin{equation}
\begin{aligned}
\nonumber \boldsymbol{y_i} &\leftarrow \boldsymbol{y_i} - \eta_l(g_i(\boldsymbol{y_i}) + \boldsymbol{c} - \boldsymbol{c_i}) \\
\nonumber c_i^+ &\leftarrow \text{(i)} \ g_i(\boldsymbol{x}), \text{or (ii)} \ \boldsymbol{c_i} - \boldsymbol{c} + \frac{1}{K\eta_l}(\boldsymbol{x} - \boldsymbol{y_i}) \\
\nonumber \boldsymbol{x} &\leftarrow \boldsymbol{x} + \frac{\eta_g}{|S|}\sum_{i \in S}(\boldsymbol{y_i} - \boldsymbol{x}) \\
\boldsymbol{c} &\leftarrow \boldsymbol{c} + \frac{1}{N}\sum_{i \in S}(\boldsymbol{c_i^+} - \boldsymbol{c_i})
\end{aligned}
\label{eq:SCAFFOLD-updates}
\end{equation}
where $\boldsymbol{y_i}$ is client update performed $K$ times, $\eta$ is the learning rate, $g_i(\boldsymbol{y_i})$ is the mini-batch gradient, and $\boldsymbol{x}$ are the server parameters. Notice there are two options to compute the local control variate $\boldsymbol{c_i}$. Option (i) makes an additional pass over the client's local data to compute the gradient at server model $\boldsymbol{x}$. Option (ii) reuses the previously computed gradients, making it the cheaper but less stable option. 

This adjustment helps ensure that the local updates are more aligned with the global model's direction, reducing the divergence caused by non-IID data distributions and other heterogeneities among clients. 


\subsection{Results}
The results for the pooled data are presented in Table \ref{table:centralized}, which highlights the testing accuracies for adopted models. Logistic regression (LR), one-layer neural network (1LNN), support vector machine (SVM) with a linear kernel achieve the highest accuracies, with SVM leading at 83.3\%. The \cite{ogier2022flamby} model refers to the LR model configured with a learning rate of 0.001 and a batch size of 4, trained over 30 epochs. 

The results for the federated setting are presented in Table \ref{tab:performance_metrics}. The models were all trained for 30 epochs using their respective best performing hyperparameter settings from centralized (pooled) tuning. We compare test accuracies (averaged over $10$ seeds) of LR, 1LNN, and SVM local models and using FedAvg, FedAdam, FedYogi and SCAFFOLD FL algorithms. The SVM local model consistently achieves the highest test accuracy across all federated algorithms, with FedYogi delivering the best performance at 0.738 ± 0.0276. 

The results in Tables \ref{table:centralized} and \ref{tab:performance_metrics} show that while LR and 1LNN perform competitively, SVM generally outperforms them in both centralized and federated settings. Notably from Table \ref{tab:performance_metrics}, federated algorithms like FedAdam and FedYogi provide improved accuracy compared to FedAvg and SCAFFOLD, demonstrating the advantages of adaptive optimization techniques in addressing the challenge of non-IID data distributions (client heterogeneity) and difficulty of hyperparameter tuning as a result. For example, in Table \ref{tab:performance_metrics},   FedAvg and SCAFFOLD perform $18\%$ worse than FedYogi for 1LNN. We believe the reason is that when the hyperparameter space becomes larger such as in 1LNN, FedAvg and SCAFFOLD become more sensitive to selection of hyperparameters. 

Further, in Table \ref{tab:performance_metrics}, we compare the FL aggregation strategies to local training on individual hospital datasets, i.e., each hospital uses only its local dataset to train a local model. We exclude Switzerland dataset from this comparison where all examples are of the same class after preprocessing. The comparison results in Table \ref{tab:performance_metrics} show that FL does improve the performance compared to local training at least for some of the hospitals. Therefore, these hospitals gain from federation without having to sacrifice privacy by data pooling or sacrifice accuracy by data anonymization. For example, for Hungary and VA hospitals, local training performs consistently worse than FedAdam and FedYogi FL algorithms for all the three learning models considered in the comparison. 


\begin{table}[!t]
\centering
\caption{Test accuracies for pooled data and various learning models.}
\begin{tabular}{l|cccccccc}
\hline
\textbf{Test} & \cite{ogier2022flamby} & LR & 1LNN & SVM \\
\textbf{Accuracy (\%)} & 78.7 & 80.5 & 82.1 & 83.3 \\
\hline
\end{tabular}
\begin{tabular}{l|cccccccc}
\hline
\textbf{Test} & NB & DT & RF & KNN \\
\textbf{Accuracy (\%)} & 78.5 & 78.1 & 80.3 & 68.6 \\
\hline
\end{tabular}
\label{table:centralized}
\end{table}

\begin{table*}[!t]
\centering
\caption{Test accuracies for local training compared to FL algorithms}
\begin{tabular}{|l|c|c|c|c|c|c|c|}
\hline
 & Cleveland & Hungary & VA & FedAvg & FedAdam & FedYogi & SCAFFOLD \\
\hline
LR & 0.667 $\pm$ 0.0831 & 0.659 $\pm$ 0.0901 & 0.608 $\pm$ 0.0616 & 0.703 $\pm$ 0.0262 & 0.72 $\pm$ 0.0191 & 0.721 $\pm$ 0.0129 & 0.703 $\pm$ 0.262 \\
\hline
1LNN & 0.728 $\pm$ 0.0462 & 0.686 $\pm$ 0.0558 & 0.648 $\pm$ 0.0742 & 0.598 $\pm$ 0.0223 & 0.726 $\pm$ 0.483 & 0.732 $\pm$ 0.329 & 0.598 $\pm$ 0.1025 \\
\hline
SVM & 0.779 $\pm$ 0.0247 & 0.628 $\pm$ 0.0873 & 0.661 $\pm$ 0.0719 & 0.731 $\pm$ 0.0223 & 0.736 $\pm$ 0.372 & 0.738 $\pm$ 0.276 & 0.731 $\pm$ 0.0223 \\
\hline
\end{tabular}
\label{tab:performance_metrics}
\end{table*}

\section{Interpretability Analysis}
In critical applications like healthcare, interpretability is essential for understanding the decision-making of predictive models. Our goal is to identify feature importance in driving model predictions. 

\begin{table*}[h]
\centering
\caption{ Mean Absolute SHAP Values for each Feature and their Importance Ranking (In Paranthesis)}
\begin{tabular}{|l|c|c|c|c|c|c|c|c|c|c|}
\hline
 & \textbf{oldpeak} & \textbf{cp} & \textbf{exang} & \textbf{sex} & \textbf{thalach} & \textbf{fbs} & \textbf{chol}  & \textbf{restecg} & \textbf{trestbps} & \textbf{age} \\
\hline
LR   &  0.57 (1)   & 0.53 (2)    & 0.51 (3)    &  0.44 (4)   & 0.37 (5)    &  0.18 (6)   & 0.16 (7)    & 0.13 (8)    & 0.03 (9)    &   0.03 (10)  \\
SVM    &  0.44(1)   &  0.44 (2)   &  0.38 (3)   & 0.36 (4)    & 0.18 (5)    & 0.14 (6)    &  0.13 (7)   &  0.04 (9)   & 0.01 (10)    &  0.08 (8)   \\
1LNN  &  0.1 (1)   & 0.07 (5)    & 0.09 (2)    & 0.05 (6)    & 0.08 (3)    & 0.02 (7)    & 0.07 (4)    &  0.02 (8)   & 0.01 (9)    &  0.0 (10)   \\
NB    &  0.11 (3)   & 0.09 (4)    &  0.15 (1)   &  0.11 (2)   &  0.06 (6)   & 0.04 (7)    & 0.06 (5)    &  0.01 (10)   &  0.03 (9)   & 0.03 (8)    \\
RF    &  0.1 (2)   &  0.17 (1)   & 0.09 (3)    & 0.08 (4)    &  0.06 (6)   & 0.02 (9)    & 0.06 (5)    & 0.01 (10)    & 0.03 (8)    & 0.04 (7)    \\
DT    &  0.0 (5)  &  0.38 (1)   &  0.05 (4)   &  0.08 (2)   &  0.0 (5)   & 0.0 (5)    & 0.06 (3)    & 0.0 (5)    &  0.0 (5)   & 0.0 (5)    \\
KNN   &   0.0 (5)  & 0.0 (5)    & 0.0 (5)    &  0.0 (5)   &  0.22 (1)   & 0.0 (5)    & 0.17 (2)    &   0.0 (5)  &  0.09 (3)   & 0.06 (4)    \\
\hline
\end{tabular}
\label{tab:mytable}
\end{table*}

\subsection{SHAP}
To address the need for interpretability, we employ SHapley Additive exPlanations (SHAP) values \cite{lundberg2017unified}, which provide a measure of feature importance in supervised learning models. SHAP is based on Shapley values from cooperative game theory which measure the marginal contribution of each feature to prediction accuracy. This is done by computing a utility (accuracy) measure for the inclusion of a feature with any other subset of features and averaging these out. In particular, the Shapley value for feature $i$ is calculated as
\begin{align}
\phi_i = \sum_{S \subseteq \{1, \ldots, p\} \setminus \{i\}} \frac{|S|!(p - |S| - 1)!}{p!} [f(S \cup i) - f(S)] 
\end{align}
where $p$ is the total number of features, $S$ is the subset (or coalition) of features excluding feature $i$, $|S|$ is the number of features in subset $S$. $f(S)$ measures the model's accuracy when only features in $S$ are input to the model. Thus, $[f(S \cup i) - f(S)]$ measures the difference in model accuracy when feature $i$ is included versus excluded from the input set $S$. This difference is computed for all possible subsets $S \subseteq \{1, \ldots, p\} \setminus \{i\}$. The $\frac{|S|!(p - |S| - 1)!}{p!}$ term serves as a weighting factor that ensures each subset's contribution is considered fairly.


Several SHAP methods adapt Shapley values to different models but share the same core principles. For example, Kernel SHAP uses kernel approximations and sampling to estimate Shapley values for complex models like neural networks through the use of a weighting kernel $\pi_{x'}$:

\begin{align}
 \pi_{x'}(z') =  \frac{M-1}{(M \ \text{choose} \ |z'|)|z'|(M-|z'|)} 
\end{align}
where $|z'|$ is the number of non-zero elements in subset $z'$, $M$ is the total number of features, and $M-|z'|$ is the number of features not in subset $z'$. The $(M \ \text{choose} \ |z'|)$ term is the binomial coefficient, which is the number of ways to sample a subset $|z'|$ from M. 

Regardless of the method, SHAP values consistently quantify each feature's contribution to predictions, improving transparency and model interpretation. 

\subsection{Importance of features}
The mean absolute SHAP values for each feature across the seven tested models are presented in Table \ref{tab:mytable}. Mean absolute SHAP values represent the average magnitude of SHAP values across all instances in the data, regardless of the direction (positive or negative) of the contribution. They indicate the overall importance of a given feature in the model, showing how much, on average, that feature influences the prediction. The values in parentheses represent the features' importance ranking based on the mean absolute SHAP values, with '1' being the highest importance and '10' being the lowest.

The results indicate that features are consistently identified as being of either high or low importance by most of the tested models, with the exceptions of Decision Tree and KNN classifiers. For example, features `oldpeak', `cp', and `exang' are ranked as high importance, while 'age', 'restecg', and 'trestbps' are low importance. `oldpeak' represents ST (the section between the end of the S wave and the beginning of the T wave) depression induced by exercise relative to rest. It measures the deviation of the ST segment on an electrocardiogram (ECG), which can indicate ischemia, which is reduced blood flow to the heart. High values of oldpeak suggest that the heart is not receiving enough oxygen during physical activity, which is a strong indicator of coronary artery disease or other heart conditions \cite{kawaji2023clinical}.

`cp' is a categorical feature that indicates the type of chest pain experienced by the patient. It is categorized into typical angina (1), atypical angina (2), non-anginal pain (3), and asymptomatic (4). The type of chest pain a patient has can be a good indicator of whether or not they have heart disease. Typical angina, for example, is a classic symptom of coronary artery disease \cite{nhlbi2024ST}. As a result, the type of chest pain can be a strong predictor of heart disease.

`exang', or exercised-induced angina, is a binary feature that indicates whether the patient experiences angina (chest pain) during exercise. If physical activity causes a patient to experience chest pain, it can be another indicator of heart disease. 

Among the features with low mean absolute SHAP values, age is notably surprising. Its general nature makes it less specific compared to clinical features directly related to heart conditions. The variability of heart disease across age groups further complicates its predictive value, with some older patients being healthy and younger patients showing symptoms. This variability may be influenced by factors like genetics, lifestyle, and pre-existing conditions, making age a less reliable indicator \cite{bots2017sex}. Similarly, 'trestbps' (resting blood pressure at admission) is also deemed low in importance. Like age, it serves as a general health indicator rather than a direct measure of heart function. High blood pressure can be a risk factor, but its predictive strength may be overshadowed by features like 'oldpeak', which directly assess heart function. Additionally, medications might lower blood pressure, masking underlying risks at admission \cite{stokes1989blood}.

The `restecg' feature represents the results of a resting electrocardiogram, categorized as: Value 0 (normal), Value 1 (ST-T wave abnormalities), and Value 2 (probable or definite left ventricular hypertrophy). Its low importance may stem from its inability to capture dynamic heart changes as effectively as stress-induced measures, such as during exercise \cite{kawaji2023clinical}.

\paragraph*{Model complexity effect on Shapley-based interpretability.} The unexpected SHAP values for Decision Trees (DT) and K-Nearest Neighbors (KNN) are due to their inherent complexity compared to simpler linear models. DTs create partitions based on features, complicating the assignment of SHAP values as each split affects the prediction path. In contrast, KNN considers instances relative to their nearest neighbors, leading to variability in SHAP values because predictions depend on different neighbor sets. This variability makes interpreting SHAP values difficult, as feature contributions are not consistently defined across instances.








\section{Conclusion}
In this work, we established a benchmark for binary classification models for heart disease using the UCI heart disease dataset, evaluated in both centralized (pooled) and federated settings, and conducted interpretability analysis using Shapley values to understand the features influencing model predictions. In the centralized setting, a linear kernel SVM achieves the highest accuracy of 83.3\%. In the federated setting, we trained various local training algorithms and model aggregation strategies. The best performing model was an SVM local trainer with testing accuracy 73.8\% for FedAdam and FedYogi aggregation strategies. Our benchmark improves on the most recent benchmark provided in \cite{ogier2022flamby} for the UCI heart disease data. It also provides reliable estimates for test accuracy since it is trained on raw data without duplication, unlike much of the existing literature. We compared the federated results with local models trained on the individual hospital datasets. In the vast majority of cases, models trained with federated aggregation outperformed the locally trained models, demonstrating the usefulness of FL. Shapley values identified oldpeak, cp, sex, and exang as the most significant features, results that coincide with medical knowledge. Future directions include providing a benchmark incorporating electrocardiogram (ECG) data to enhance performance in both centralized and federated settings, and extending interpretability analysis to deep models trained on ECG data.

\bibliographystyle{IEEEtran}
\bibliography{IEEEabrv, mybibfile}

\begin{thebibliography}{10}
\providecommand{\url}[1]{#1}
\csname url@samestyle\endcsname
\providecommand{\newblock}{\relax}
\providecommand{\bibinfo}[2]{#2}
\providecommand{\BIBentrySTDinterwordspacing}{\spaceskip=0pt\relax}
\providecommand{\BIBentryALTinterwordstretchfactor}{4}
\providecommand{\BIBentryALTinterwordspacing}{\spaceskip=\fontdimen2\font plus
\BIBentryALTinterwordstretchfactor\fontdimen3\font minus
  \fontdimen4\font\relax}
\providecommand{\BIBforeignlanguage}[2]{{%
\expandafter\ifx\csname l@#1\endcsname\relax
\typeout{** WARNING: IEEEtran.bst: No hyphenation pattern has been}%
\typeout{** loaded for the language `#1'. Using the pattern for}%
\typeout{** the default language instead.}%
\else
\language=\csname l@#1\endcsname
\fi
#2}}
\providecommand{\BIBdecl}{\relax}
\BIBdecl

\bibitem{UCI_heart}
A.~Janosi, W.~Steinbrunn, M.~Pfisterer, and R.~Detrano, ``{Heart Disease},''
  UCI Machine Learning Repository, 1988, {DOI}:
  https://doi.org/10.24432/C52P4X.

\bibitem{puhan2023digital}
B.~Puhan and S.~Gupta, ``Digital health rules and regulations: an overview,''
  \emph{CSI Transactions on ICT}, vol.~11, no.~1, pp. 97--102, 2023.

\bibitem{sweeney2002k}
L.~Sweeney, ``k-anonymity: {A} model for protecting privacy,''
  \emph{International journal of uncertainty, fuzziness and knowledge-based
  systems}, vol.~10, no.~05, pp. 557--570, 2002.

\bibitem{rieke2020future}
N.~Rieke, J.~Hancox, W.~Li, F.~Milletari, H.~R. Roth, S.~Albarqouni, S.~Bakas,
  M.~N. Galtier, B.~A. Landman, K.~Maier-Hein \emph{et~al.}, ``The future of
  digital health with federated learning,'' \emph{NPJ digital medicine},
  vol.~3, no.~1, p. 119, 2020.

\bibitem{xu2021federated}
J.~Xu, B.~S. Glicksberg, C.~Su, P.~Walker, J.~Bian, and F.~Wang, ``Federated
  learning for healthcare informatics,'' \emph{Journal of Healthcare
  Informatics Research}, vol.~5, pp. 1--19, 2021.

\bibitem{nguyen2022federated}
D.~C. Nguyen, Q.-V. Pham, P.~N. Pathirana, M.~Ding, A.~Seneviratne, Z.~Lin,
  O.~Dobre, and W.-J. Hwang, ``Federated learning for smart healthcare: A
  survey,'' \emph{ACM Computing Surveys (CSUR)}, vol.~55, no.~3, pp. 1--37,
  2022.

\bibitem{ogier2022flamby}
J.~Ogier~du Terrail, S.-S. Ayed, E.~Cyffers, F.~Grimberg, C.~He, R.~Loeb,
  P.~Mangold, T.~Marchand, O.~Marfoq, E.~Mushtaq \emph{et~al.}, ``Flamby:
  Datasets and benchmarks for cross-silo federated learning in realistic
  healthcare settings,'' \emph{Advances in Neural Information Processing
  Systems}, vol.~35, pp. 5315--5334, 2022.

\bibitem{Konecny2016federated}
\BIBentryALTinterwordspacing
J.~Konečný, H.~B. McMahan \emph{et~al.}, ``Federated learning: {S}trategies
  for improving communication efficiency,'' in \emph{NIPS Workshop on Private
  Multi-Party Machine Learning}, 2016. [Online]. Available:
  \url{https://arxiv.org/abs/1610.05492}
\BIBentrySTDinterwordspacing

\bibitem{mcmahan2017communication}
B.~McMahan, E.~Moore, D.~Ramage, S.~Hampson, and B.~A. y~Arcas,
  ``Communication-efficient learning of deep networks from decentralized
  data,'' in \emph{Artificial intelligence and statistics}.\hskip 1em plus
  0.5em minus 0.4em\relax PMLR, 2017, pp. 1273--1282.

\bibitem{kairouz2021advances}
P.~Kairouz, H.~B. McMahan, B.~Avent, A.~Bellet, M.~Bennis, A.~N. Bhagoji,
  K.~Bonawitz, Z.~Charles, G.~Cormode, R.~Cummings \emph{et~al.}, ``Advances
  and open problems in federated learning,'' \emph{Foundations and
  trends{\textregistered} in machine learning}, vol.~14, no. 1--2, pp. 1--210,
  2021.

\bibitem{du2021collaborative}
J.~O. Du~Terrail, A.~Leopold, C.~Joly, C.~Beguier, M.~Andreux, C.~Maussion,
  B.~Schmauch, E.~W. Tramel, E.~Bendjebbar, M.~Zaslavskiy \emph{et~al.},
  ``Collaborative federated learning behind hospitals’ firewalls for
  predicting histological response to neoadjuvant chemotherapy in
  triple-negative breast cancer,'' \emph{medRxiv}, pp. 2021--10, 2021.

\bibitem{li2020using}
W.~T. Li, J.~Ma, N.~Shende, G.~Castaneda, J.~Chakladar, J.~C. Tsai, L.~Apostol,
  C.~O. Honda, J.~Xu, L.~M. Wong \emph{et~al.}, ``Using machine learning of
  clinical data to diagnose covid-19: a systematic review and meta-analysis,''
  \emph{BMC medical informatics and decision making}, vol.~20, pp. 1--13, 2020.

\bibitem{pati2021federated}
S.~Pati, U.~Baid, M.~Zenk, B.~Edwards, M.~Sheller, G.~A. Reina, P.~Foley,
  A.~Gruzdev, J.~Martin, S.~Albarqouni \emph{et~al.}, ``The federated tumor
  segmentation (fets) challenge,'' \emph{arXiv preprint arXiv:2105.05874},
  2021.

\bibitem{karimireddy2020scaffold}
S.~P. Karimireddy, S.~Kale, M.~Mohri, S.~Reddi, S.~Stich, and A.~T. Suresh,
  ``Scaffold: Stochastic controlled averaging for federated learning,'' in
  \emph{International conference on machine learning}.\hskip 1em plus 0.5em
  minus 0.4em\relax PMLR, 2020, pp. 5132--5143.

\bibitem{li2020federated}
T.~Li, A.~K. Sahu, M.~Zaheer, M.~Sanjabi, A.~Talwalkar, and V.~Smith,
  ``Federated optimization in heterogeneous networks,'' \emph{Proceedings of
  Machine learning and systems}, vol.~2, pp. 429--450, 2020.

\bibitem{reddi2020adaptive}
S.~Reddi, Z.~Charles, M.~Zaheer, Z.~Garrett, K.~Rush, J.~Kone{\v{c}}n{\`y},
  S.~Kumar, and H.~B. McMahan, ``Adaptive federated optimization,'' \emph{arXiv
  preprint arXiv:2003.00295}, 2020.

\bibitem{cho2021personalized}
Y.~J. Cho, J.~Wang, T.~Chiruvolu, and G.~Joshi, ``Personalized federated
  learning for heterogeneous clients with clustered knowledge transfer,''
  \emph{arXiv preprint arXiv:2109.08119}, 2021.

\bibitem{fallah2020personalized}
A.~Fallah, A.~Mokhtari, and A.~Ozdaglar, ``Personalized federated learning with
  theoretical guarantees: A model-agnostic meta-learning approach,''
  \emph{Advances in Neural Information Processing Systems}, vol.~33, pp.
  3557--3568, 2020.

\bibitem{ghosh2020efficient}
A.~Ghosh, J.~Chung, D.~Yin, and K.~Ramchandran, ``An efficient framework for
  clustered federated learning,'' in \emph{the 33rd Conference on Advances in
  Neural Information Processing Systems (NeurIPS)}, 2020, pp. 19\,586--19\,597.

\bibitem{jiang2019improving}
Y.~Jiang, J.~Konečný, K.~Rush, and S.~Kannan, ``{Improving federated learning
  personalization via model agnostic meta learning},'' \emph{arXiv preprint
  arXiv:1909.12488}, 2019.

\bibitem{nafea2022proportional}
M.~Nafea, E.~Shin, and A.~Yener, ``Proportional fair clustered federated
  learning,'' in \emph{IEEE International Symposium on Information Theory
  (ISIT)}.\hskip 1em plus 0.5em minus 0.4em\relax IEEE, 2022.

\bibitem{sattler2019clustered}
F.~Sattler, K.-R. Müller, and W.~Samek, ``{Clustered federated learning:
  {M}odel-agnostic distributed multi-task optimization under privacy
  constraints},'' \emph{arXiv preprint arXiv:1910.01991}, 2019.

\bibitem{smith2017federated}
V.~Smith, C.-K. Chiang, M.~Sanjabi, and A.~S. Talwalkar, ``{Federated
  multi-Task learning},'' in \emph{Advances in Neural Information Processing
  Systems}, vol.~30, 2017.

\bibitem{hsu2019measuring}
T.-M.~H. Hsu, H.~Qi, and M.~Brown, ``Measuring the effects of non-identical
  data distribution for federated visual classification,'' \emph{arXiv preprint
  arXiv:1909.06335}, 2019.

\bibitem{datta2016algorithmic}
A.~Datta, S.~Sen, and Y.~Zick, ``Algorithmic transparency via quantitative
  input influence: Theory and experiments with learning systems,'' in
  \emph{2016 IEEE symposium on security and privacy (SP)}.\hskip 1em plus 0.5em
  minus 0.4em\relax IEEE, 2016, pp. 598--617.

\bibitem{doshi2017towards}
F.~Doshi-Velez and B.~Kim, ``Towards a rigorous science of interpretable
  machine learning,'' \emph{arXiv preprint arXiv:1702.08608}, 2017.

\bibitem{arrieta2020explainable}
A.~B. Arrieta, N.~D{\'\i}az-Rodr{\'\i}guez, J.~Del~Ser, A.~Bennetot, S.~Tabik,
  A.~Barbado, S.~Garc{\'\i}a, S.~Gil-L{\'o}pez, D.~Molina, R.~Benjamins
  \emph{et~al.}, ``Explainable artificial intelligence (xai): Concepts,
  taxonomies, opportunities and challenges toward responsible ai,''
  \emph{Information fusion}, vol.~58, pp. 82--115, 2020.

\bibitem{shapley1953value}
L.~S. Shapley \emph{et~al.}, \emph{A value for n-person games}.\hskip 1em plus
  0.5em minus 0.4em\relax Princeton University Press Princeton, 1953.

\bibitem{lundberg2017unified}
S.~M. Lundberg and S.-I. Lee, ``A unified approach to interpreting model
  predictions,'' \emph{Advances in neural information processing systems},
  vol.~30, 2017.

\bibitem{ncbi2024ST}
\BIBentryALTinterwordspacing
N.~C. for Biotechnology~Information, ``St segment,'' 2024. [Online]. Available:
  \url{https://www.ncbi.nlm.nih.gov/books/NBK459364/}
\BIBentrySTDinterwordspacing

\bibitem{kawaji2023clinical}
T.~Kawaji, Y.~Hamatani, M.~Kato, T.~Yokomatsu, S.~Miki, M.~Abe, and M.~Akao,
  ``Clinical significance of st-segment depression during atrial fibrillation
  rhythm for subsequent heart failure events,'' \emph{European Heart Journal
  Open}, vol.~3, no.~3, p. oead060, 2023.

\bibitem{nhlbi2024ST}
\BIBentryALTinterwordspacing
NHLBI, ``Angina (chest pain) causes and risk factors,'' 2024. [Online].
  Available:
  \url{https://www.nhlbi.nih.gov/health/angina/causes#:~:text=The%20main%20lifestyle%20risk%20factors,Medical%20procedures}
\BIBentrySTDinterwordspacing

\bibitem{ahsan2022machine}
M.~M. Ahsan and Z.~Siddique, ``Machine learning-based heart disease diagnosis:
  A systematic literature review,'' \emph{Artificial Intelligence in Medicine},
  vol. 128, p. 102289, 2022.

\bibitem{aljanabi2018machine}
M.~Aljanabi, M.~H. Qutqut, and M.~Hijjawi, ``Machine learning classification
  techniques for heart disease prediction: a review,'' \emph{International
  Journal of Engineering \& Technology}, vol.~7, no.~4, pp. 5373--5379, 2018.

\bibitem{azmi2022systematic}
J.~Azmi, M.~Arif, M.~T. Nafis, M.~A. Alam, S.~Tanweer, and G.~Wang, ``A
  systematic review on machine learning approaches for cardiovascular disease
  prediction using medical big data,'' \emph{Medical engineering \& physics},
  vol. 105, p. 103825, 2022.

\bibitem{kaggle_heart}
{Kaggle}, ``Heart disease dataset,''
  \url{https://www.kaggle.com/datasets/johnsmith88/heart-disease-dataset/data},
  n.d., accessed: February 20, 2024.

\bibitem{prasher2023evaluation}
S.~Prasher, L.~Nelson, and S.~Hariharan, ``Evaluation of machine learning
  algorithms for heart disease prediction in healthcare,'' in \emph{2023
  International Conference on Innovations in Engineering and Technology
  (ICIET)}.\hskip 1em plus 0.5em minus 0.4em\relax IEEE, 2023, pp. 1--4.

\bibitem{ali2021heart}
M.~M. Ali, B.~K. Paul, K.~Ahmed, F.~M. Bui, J.~M. Quinn, and M.~A. Moni,
  ``Heart disease prediction using supervised machine learning algorithms:
  Performance analysis and comparison,'' \emph{Computers in Biology and
  Medicine}, vol. 136, p. 104672, 2021.

\bibitem{ahmad2022efficient}
G.~N. Ahmad, H.~Fatima, S.~Ullah, A.~S. Saidi \emph{et~al.}, ``Efficient
  medical diagnosis of human heart diseases using machine learning techniques
  with and without gridsearchcv,'' \emph{IEEE Access}, vol.~10, pp.
  80\,151--80\,173, 2022.

\bibitem{ahamad2023influence}
G.~N. Ahamad, Shafiullah, H.~Fatima, Imdadullah, S.~Zakariya, M.~Abbas, M.~S.
  Alqahtani, and M.~Usman, ``Influence of optimal hyperparameters on the
  performance of machine learning algorithms for predicting heart disease,''
  \emph{Processes}, vol.~11, no.~3, p. 734, 2023.

\bibitem{ayano2022interpretable}
Y.~M. Ayano, F.~Schwenker, B.~D. Dufera, and T.~G. Debelee, ``Interpretable
  machine learning techniques in ecg-based heart disease classification: a
  systematic review,'' \emph{Diagnostics}, vol.~13, no.~1, p. 111, 2022.

\bibitem{anand2022explainable}
A.~Anand, T.~Kadian, M.~K. Shetty, and A.~Gupta, ``Explainable ai decision
  model for ecg data of cardiac disorders,'' \emph{Biomedical Signal Processing
  and Control}, vol.~75, p. 103584, 2022.

\bibitem{goodfellow2018towards}
S.~D. Goodfellow, A.~Goodwin, R.~Greer, P.~C. Laussen, M.~Mazwi, and D.~Eytan,
  ``Towards understanding ecg rhythm classification using convolutional neural
  networks and attention mappings,'' in \emph{Machine learning for healthcare
  conference}.\hskip 1em plus 0.5em minus 0.4em\relax PMLR, 2018, pp. 83--101.

\bibitem{zhou2016learning}
B.~Zhou, A.~Khosla, A.~Lapedriza, A.~Oliva, and A.~Torralba, ``Learning deep
  features for discriminative localization,'' in \emph{Proceedings of the IEEE
  conference on computer vision and pattern recognition}, 2016, pp. 2921--2929.

\bibitem{wang2021interpretable}
K.~Wang, J.~Tian, C.~Zheng, H.~Yang, J.~Ren, Y.~Liu, Q.~Han, and Y.~Zhang,
  ``Interpretable prediction of 3-year all-cause mortality in patients with
  heart failure caused by coronary heart disease based on machine learning and
  shap,'' \emph{Computers in biology and medicine}, vol. 137, p. 104813, 2021.

\bibitem{sethi2024explainable}
A.~Sethi, S.~Dharmavaram, and S.~Somasundaram, ``Explainable artificial
  intelligence (xai) approach to heart disease prediction,'' in \emph{2024 3rd
  International Conference on Artificial Intelligence For Internet of Things
  (AIIoT)}.\hskip 1em plus 0.5em minus 0.4em\relax IEEE, 2024, pp. 1--6.

\bibitem{kingma2014adam}
D.~P. Kingma and J.~Ba, ``Adam: A method for stochastic optimization,''
  \emph{arXiv preprint arXiv:1412.6980}, 2014.

\bibitem{zaheer2018adaptive}
M.~Zaheer, S.~Reddi, D.~Sachan, S.~Kale, and S.~Kumar, ``Adaptive methods for
  nonconvex optimization,'' \emph{Advances in neural information processing
  systems}, vol.~31, 2018.

\bibitem{bots2017sex}
S.~H. Bots, S.~A. Peters, and M.~Woodward, ``Sex differences in coronary heart
  disease and stroke mortality: a global assessment of the effect of ageing
  between 1980 and 2010,'' \emph{BMJ global health}, vol.~2, no.~2, p. e000298,
  2017.

\bibitem{stokes1989blood}
J.~Stokes~3rd, W.~B. Kannel, P.~A. Wolf, R.~B. D'Agostino, and L.~A. Cupples,
  ``Blood pressure as a risk factor for cardiovascular disease. the framingham
  study--30 years of follow-up.'' \emph{Hypertension}, vol.~13, no.
  5\_supplement, p. I13, 1989.

\end{thebibliography}

\end{document}